# Supervised learning sets benchmark for robust spike detection from calcium imaging signals

Lucas Theis[1,2,*], Philipp Berens[$,*,1,2,3,4,5], Emmanouil Froudarakis[4], Jacob Reimer[4], Miroslav Román Rosón[1,5], Tom Baden[1,3,5], Thomas Euler[1,3,5], Andreas Tolias[3,4,6], Matthias Bethge[1,2,3,$]

[1] Centre for Integrative Neuroscience, University of Tübingen, Tübingen, Germany
[2] Institute of Theoretical Physics, University of Tübingen, Tübingen, Germany
[3] Bernstein Center for Computational Neuroscience, University of Tübingen, Tübingen, Germany
[4] Department of Neuroscience, Baylor College of Medicine, Houston, USA
[5] Institute for Ophthalmic Research, University of Tübingen, Tübingen, Germany
[6] Department of Computational and Applied Mathematics, Rice University, Houston, USA

* These authors contributed equally to this work.

$ To whom correspondence should be addressed:
Philipp Berens, philipp.berens@uni-tuebingen.de
Matthias Bethge, matthias.bethge@uni-tuebingen.de

## Summary

A fundamental challenge in calcium imaging has been to infer the timing of action potentials from the measured noisy calcium fluorescence traces. We systematically evaluate a range of spike inference algorithms on a large benchmark dataset recorded from varying neural tissue (V1 and retina) using different calcium indicators (OGB-1 and GCamp6). We show that a new algorithm based on supervised learning in flexible probabilistic models outperforms all previously published techniques, setting a new standard for spike inference from calcium signals. Importantly, it performs better than other algorithms even on datasets not seen during training. Future data acquired in new experimental conditions can easily be used to further improve its spike prediction accuracy and generalization performance. Finally, we show that comparing algorithms on artificial data is not informative about performance on real population imaging data, suggesting that a benchmark dataset may greatly facilitate future algorithmic developments.

# Introduction

Over the past two decades, two-photon imaging has become one of the most widely used techniques for studying information processing in neural populations in vivo (Denk et al., 1990; Kerr and Denk, 2008). Typically, a calcium indicator such as the synthetic dye Oregon green BAPTA-1 (OGB-1) (Stosiek et al., 2003) or the genetically encoded GCamp6 (Chen et al., 2013) is used to image a large fraction of cells in a neural tissue. Individual action potentials lead to a fast rise in fluorescence, followed by a slow decay with a time constant of several hundred milliseconds (Chen et al., 2013; Kerr et al., 2005). Commonly, neural population activity from dozens or hundreds of cells is imaged using relatively slow scanning speeds (<15 Hz), but novel fast scanning methods (Cotton et al., 2013; Grewe et al., 2010) (up to several 100 Hz) have opened additional opportunities for studying neural population activity at increased temporal resolution.

A fundamental challenge has been to infer the timing of action potentials from the measured noisy calcium fluorescence traces. To solve this problem of spike inference, several different approaches have been proposed, including template-matching (Greenberg et al., 2008; Grewe et al., 2010; Oñativia et al., 2013) and deconvolution (Park et al., 2013; Pnevmatikakis et al., 2013, 2014; Vogelstein et al., 2009, 2010; Yaksi and Friedrich, 2006). These methods have in common that they assume a forward generative model of calcium signal generation which is then inverted to infer spike times. A crucial shortcoming of this approach is that the forward models make strong a-priori assumptions about the shape of the calcium fluorescence signal induced by a single spike and the statistics of the noise. Alternatively, simple supervised learning techniques, which learn their parameters from data, have been used to infer spikes from calcium signals (Sasaki et al., 2008).

However, it is currently not known which approach is most successful at inferring spikes under experimental conditions, as a detailed quantitative comparison of different algorithms on large datasets of *in vitro* and *in vivo* population imaging data has been lacking. Rather, most published algorithms have only been evaluated on relatively small experimental datasets using different performance measures. In addition, the question of how well we can reconstruct the spikes of neurons given calcium measurements has been studied theoretically or using simulated datasets (Lütcke et al., 2013; Wilt et al., 2013). While such studies offer the advantage that many model parameters are under the control of the investigator, they do not answer the question of how well we can reconstruct spikes from actual measurements.

Here, we advocate a data-driven approach based on flexible probabilistic models to infer spikes from calcium fluorescence traces. We collected a large benchmark dataset including simultaneous measurements of spikes and calcium signals in primary visual cortex and the retina of mice using OGB-1 and GCamp6 as calcium indicators. We systematically evaluate a range of spike inference algorithms and show that supervised learning in flexible probabilistic models outperforms all previously published techniques, setting a new standard for spike inference from calcium signals.

# Results

**A flexible probabilistic model for spike inference**

We propose to model the probabilistic relationship between a segment of the fluorescence trace $x_t$ and the number of spikes $k_t$ in a small time bin, assuming they are Poisson distributed with rate $\lambda(x_t)$:

$$p(k_t \mid x_t) = \frac{\lambda(x_t)^k}{k!} e^{-\lambda(x_t)}.$$

Instead of relying on a specific forward model, we modeled the firing rate $\lambda(x_t)$ using a recently introduced extension of generalized linear models, the factored spike-triggered mixture (STM) model (Theis et al., 2013) (Fig. 1a; see Methods):

$$\lambda_{\text{STM}}(x_t) = \sum_{k=1}^{K} \exp\left(\sum_{m=1}^{M} \beta_{km}(u_m^\top x_t)^2 + w_k^\top x_t + b_k\right).$$

We train this model on simultaneous recordings of spikes and calcium traces to learn a set of $K$ linear features $w_k$ and $M$ quadratic features $u_m$ ('supervised learning'), which are predictive of the occurrence of spikes in the fluorescence trace. Importantly, this model is sufficiently flexible to capture non-linear relationships between fluorescence traces and spikes, but at the same time is sufficiently restricted to avoid overfitting when little data is available. Below we will evaluate whether this model is too simple or already more complex than necessary by comparing its performance to that of multi-layer neural networks and simple LNP-type models.

*Fig. 1: Spike inference from calcium measurements*

Using a probabilistic model in this way not only provides us with an estimate of the expected firing rate, $\lambda(x_t)$, but also with access to a distribution over spike counts, as fully Bayesian methods do (Pnevmatikakis et al., 2013, 2014; Vogelstein et al., 2009). This allows us to estimate the uncertainty in the predictions and to generate example spikes trains without spending considerable computational resources.

**Benchmarking spike inference algorithms on experimental data**

To quantitatively evaluate different spike inference approaches including our model, we acquired a large benchmark dataset with a total of 75 traces of 67 neurons, in which we simultaneously recorded calcium signals and spikes (Fig. 1b; in total ~ 89,000 spikes). These cells were recorded with different scanning methods, different calcium indicators and at different sampling rates (see *Table 1* and *Methods*): Dataset 1 consisted of 16 neurons recorded *in-vivo* in V1 of anesthetized mice using fast 3D AOD-based imaging (Cotton et al., 2013) at ~320 Hz with OGB-1 as indicator. Dataset 2 consisted of 31 neurons recorded *in-vivo* in mouse V1 using line scanning at ~12 Hz with OGB-1 as indicator. Dataset 3 consisted of 19 segments recorded from 11 neurons in-vivo in mouse V1 using the genetic calcium indicator GCamp6s with a resonance scanner at ~59 Hz. Finally, dataset 4 consisted of 9 retinal ganglion cells recorded *in-vitro* at ~8 Hz using line-scanning with OGB-1 as indicator (Briggman and Euler, 2011). We resampled the calcium traces from all four datasets to a common resolution of 100 Hz. All datasets were acquired at a zoom factor commonly used in population imaging such that the signal quality should match well that commonly encountered in these preparations (see *Table 1*).

We compared the performance of our algorithm (*STM*) to that of algorithms representative of the different approaches (see *Table 2* and *Methods*), including simple deconvolution (*YF06*, Yaksi and Friedrich, 2006), MAP (*VP10*, known as 'fast-oopsi', Vogelstein et al., 2010) and Bayesian inference (*PP14*, Pnevmatikakis et al., 2014; *VP09*, Vogelstein et al., 2009) in generative models, template-matching by finite rate of innovation (*OD13*, Oñativia et al., 2013) and supervised learning using a support vector machine (*SI08*, Sasaki et al., 2008). To provide a baseline level of performance, we evaluated how closely the calcium trace followed the spike train without any further processing (*raw*).

We focus on two measures of spike reconstruction performance to provide a quantitative evaluation of the different techniques: (i) the correlation between the original and the reconstructed spike train and (ii) the information gained about the spike train based on the calcium signal (see *Methods*). For completeness, we computed (iii) the area under the ROC curve (AUC), a threshold independent measure of the spike detection accuracy, which has repeatedly been used in the literature. We focus on correlation and information gain, however, since the AUC score is insensitive to changes in the relative height of different parts of the spike density function (e.g. high rates could be consistently overestimated compared to low rates; for a more technical discussion, see *Methods*).

To provide a fair comparison between the different algorithms, we evaluated their performance using leave-one-out cross-validation: we estimated the parameters of the algorithms on all but one cell from a dataset and tested them on the one remaining neuron, repeating this procedure for each neuron in the dataset (cross-validation; see *Methods*). For the algorithms based on generative models, we selected the hyperparameters during cross-validation (VP10, VP09) or using a sampling based approach (PP14; see *Methods*).

**Supervised learning sets benchmark**
We found that the spike density function predicted by our algorithm matched the true spike train closely, for cells from each dataset including both indicators OGB-1 and GCamp6 (Fig. 1c-f). The other tested algorithms generally showed worse prediction performance: For example, *YF06* typically resulted in very noisy estimates of the spike density function (Fig. 1c-f) and both *VP10* and *PP14* frequently missed single spikes (Fig. 1d-f, marked by asterisk) and had difficulties modeling the dynamics of the GCamp6 indicator (Fig. 1e).

Figure 2: Quantitative evaluation of spike inference performance

A quantitative comparison revealed that our STM method reconstructed the true spike trains better than all its competitors, yielding a consistently higher correlation and information gain for all four datasets (Fig. 2a, b; evaluated at 25 Hz; for statistics, see figure). The median improvement in correlation across all recordings achieved by the STM over its two closest competitors was 0.12 (0.07-0.14; median and bootstrapped 95%-confidence interval, N=75) for *SI08* – the other supervised learning approach based on SVMs – and 0.1 (0.08-0.13) for *PP14* – the Bayesian inference in a generative model – yielding a median improvement of 33% and 32%, respectively. Similarly, the STM explained 6.8 (5.0-7.7; *SI08*) and 9.6 (8.1-12.1; PP14) percent points more marginal entropy (measured by the relative information gain). When evaluated with respect to AUC, the performance of these two algorithms was about as good as that of the STM model (Suppl. Fig. 1), yielding a median difference in AUC of -0.01 (-0.02-0.01) and 0.01 (-0.01-0.02). This is because the AUC is the least sensitive of the three measures, as discussed above.

*Figure 3: Spike inference performance as a function of frequency*

The performance advantage of our STM algorithm was not restricted to a particular choice of sampling rate: it performed better than all other methods for a wide range of sampling rates between 2 and 100 Hz, corresponding to time bins between 10 and several hundreds of milliseconds (Fig. 3). Its performance advantage was particularly large for high sampling rates (Fig. 3; also Suppl. Fig. 2), suggesting that the timing accuracy of our method is superior to that of other methods. Interestingly, *VP10* ('fast-oopsi') performed similar to our method for low sampling rates, but its performance deteriorated consistently on all datasets to the performance level of VF06 with increasing sampling rates (Fig. 3).

*Figure 4: Evaluating model complexity*

The performance of the STM model could not be further improved using a more flexible multilayer neural network for modeling the non-linear rate function $\lambda_t$ (Fig. 4 and Suppl. Fig. 3). To test this, we replaced the STM model by a neural network with two hidden layer, but found that this change resulted in only marginal performance improvement (Fig. 4). In addition, we tested whether a much simpler linear-nonlinear model would suffice to model $\lambda_t$. We found that the STM model performed significantly better than the simple LNP model (Fig. 4 and Suppl. Fig. 3). Therefore, the choice of the STM for $\lambda_t$ seems to provide a good compromise between flexibility of the model structure and generalization performance.

**Does the performance generalize to new datasets?**
Remarkably, the STM model was able to generalize to new data sets that were recorded under different conditions than the data used for training. To test this, we trained the algorithms on three of the datasets and evaluated it on the remaining one (Fig. 5a). This setup mimics the situation where no simultaneous spike-calcium recordings are available for a new preparation, scanning method or calcium indicator.

*Figure 5: Testing generalization performance*

The STM algorithm still showed better performance compared to all other algorithms (Fig. 5b-c and Suppl. Fig. 4), including superior performance on the GCamp6-dataset when trained solely on the three OGB-datasets (Fig. 5b-c). This indicates that the algorithm may be directly applied on novel datasets without need for further training (*see Discussion*).

Already a small training set of less than 10 cells was sufficient to achieve good performance for the STM model (Suppl. Fig. 5). We tested the prediction performance of the STM with training sets of various sizes and found that it saturated between 5 and 10 cells for all datasets, arguing that a few cells may suffice to directly adapt the model to new datasets.

**Comparisons on artificial data**
Surprisingly, the performance of the algorithms on simulated data was not predictive of the performance of the algorithms on the real datasets (Fig. 6). To test this, we simulated data from a simple biophysical model of calcium fluorescence generation (Fig. 6a, see *Methods*, Vogelstein et al., 2009). We then applied the same cross-validation procedure as before to evaluate the performance of the algorithms (Fig. 6b). Not surprisingly, we found that all algorithms based on this or a similar generative model (*PP13*, *VP10*, *YF06*) performed remarkably well. Interestingly, even the algorithms that performed worse than the baseline model for the real data (*OD13*, *VP09*) showed good performance on the artificial data. The STM model was among the top-performing algorithms, in contrast to the other supervised learning algorithm (*SI08*). A direct comparison of the performance on the simulated dataset and the experimental data clearly illustrates that the former is not a good predictor of the latter (Fig. 6c).

*Figure 6: Evaluating algorithms on artificial data*

## Discussion

We introduced a new flexible probabilistic model for inferring spikes from calcium traces based on supervised learning. We showed that this model performs better than all previously published algorithms for this problem, for a wide range of recording conditions including OGB-1 and GCamp6 as calcium indicators, different scanning techniques, neural tissues, and with respect to different metrics. Interestingly, two of the three best algorithms rely on supervised learning to infer the relationship between calcium signal and spikes, suggesting that a data-driven approach offers distinct advantages over approaches based on strong prior suppositions about the relationship between the two signals.

The superior performance of our algorithm carried over to new datasets not seen during training, promising good spike inference performance even under experimental conditions where no simultaneous recordings are available. This is crucial, as this is often considered an important advantage of algorithms based on generative models. However, for new experimental conditions, the performance of this latter class of algorithms is by no means guaranteed and needs to be evaluated on a dataset with simultaneous recordings as well. In particular, if such an evaluation reveals poor performance, e.g. because the assumed generative model does not match the structure of the dataset at hand (as seen e.g. with the GCamp6 data; Fig. 1e and 2), the only way to improve the algorithm would be to adapt the generative model and modify the inference procedures accordingly, which may or may not be straightforward. In contrast, any simultaneous data collected in the future can be readily used to retrain our supervised algorithm and further improve its spike prediction and generalization performance. In fact, our choice of the spike triggered mixture model for estimating spikes from calcium traces is motivated by its ability to automatically switch between different sub-models whenever the statistics of the data changes (Theis et al., 2013).

Our evaluation further shows that good spike inference performance on model data by no means guarantees good performance on real population imaging data (Fig. 6c). We believe theoretical model based studies (Lütcke et al., 2013; Wilt et al., 2013) will remain useful to systematically explore how performance depends on model parameters, such as noise level or violations of the generative model, but will need to be followed up by systematic quantitative benchmark comparisons on datasets such as provided here.

Our proposed method is solely concerned with the problem of spike inference, and does not infer the regions of interests (ROIs) from observed data. Rather, we assume that these are obtained by the experimenter through other semi-automatic or automatic techniques. Recently, several methods have been proposed to jointly infer ROIs and spikes (Diego and Hamprecht, 2014; Maruyama et al., 2014; Pnevmatikakis et al., 2014). These methods have the benefit that they exploit the full spatio-temporal structure of the problem of spike inference in calcium imaging and offer an unbiased approach for ROI placement. Since ROIs can also be placed using supervised learning (Valmianski et al., 2010), it should be feasible to develop supervised paradigms for simultaneous ROI placement and spike inference or combinations of unsupervised and supervised methods.

## A challenge

We presented the first quantitative benchmarking approach to evaluating spike inference algorithms on a large dataset of population imaging data. Our proposed algorithm performed much better than all previously suggested ones – we are now looking for even better algorithms! To find them, we have set up a competition with Kaggle, where a part of our dataset is available for download. Based on this dataset, the community is invited to build the best algorithm possible over the coming months, and submit it to be evaluated. We would like to explore the limits of spike inference, and make the best-performing tools available to everybody. We believe that this benchmarking approach which is already used successfully in machine learning and related fields to drive new algorithmic developments can also be an important catalyst for improvements on various computational problems in neuroscience, from systems identification to neuron reconstruction.

# Methods

## Datasets

*Primary visual cortex (V1) – OGB-1*

We recorded calcium traces from neural populations in layer 2/3 of anesthetized wild type mice (male C57CL/6J, age: p40–p60) using a custom-built two-photon microscope using previously described methods (Cotton et al., 2013; Froudarakis et al., 2014). Briefly, the temperature of the mouse was maintained between 36.5 °C and 37.5 °C throughout the experiment using a homeothermic blanket system (Harvard Instruments). While recording we either provided no visual stimulation, moving gratings, or natural and phase scrambled movies as previously described (Froudarakis et al., 2014). A ~1 mm craniotomy was performed over the primary visual cortex of the mouse. The details of surgical techniques and anesthesia protocol have been described elsewhere (Cotton et al., 2013). We then used bolus-loaded Oregon green BAPTA-1 (OGB-1, Invitrogen) as calcium indicator and the injections were performed by using a continuous-pulse low pressure protocol with a glass micropipette to inject ~300 μm below the surface of the cortex. The cortical window was sealed using a glass coverslip. After allowing 1h for the dye uptake we recorded calcium traces using a custom-built two-photon microscope equipped with a Chameleon Ti-sapphire laser (Coherent) tuned at 800 nm and a 20×, 1.0 NA Olympus objective. Scanning was controlled by either a set of galvanometric mirrors (*Galvo*) or a custom-built acousto-optic deflector system (*AODs*) (Cotton et al., 2013). The average power output of the objective was kept < 50 mW for galvanometric scanning and 120 mW for AODs. Calcium activity was typically sampled at ~12 Hz with the galvanometric mirrors and at ~320 Hz with the AODs. The field of view was typically 200x200x100μm and 250x250μm for AODs and galvanometric imaging, respectively, imaging dozens to hundreds of neurons simultaneously(Cotton et al., 2013). To perform simultaneous loose-patch and two-photon calcium imaging recordings, we used glass pipettes with 5–7 MΩ resistance filled with Alexa Fluor 594 (Invitrogen) for targeted two-photon-guided loose cell patching of single cells. Spike times were extracted by thresholding. All procedures performed on mice were conducted in accordance with the ethical guidelines of the National Institutes of Health and were approved by the Baylor College of Medicine IACUC.

*Primary visual cortex (V1) – GCaMP6*

We recorded calcium traces from neural populations in layer 2/3 of isoflurane-anesthetized wild type mice (male C57CL/6J, age: p70-p80) using a resonant scanning microscope (ThorLabs). Surgical procedures were similar to those described in Reimer et al (2014). Briefly, mice were initially injected with approximately 1uL of AAV1.Syn.GCamp6s.WPRE.SV40 (University of Pennsylvania Vector Core) through a burr hole. The injection was performed with the pipette at a steep (~60 deg) angle, in order to infect cells in the cortex lateral to the injection site under an untouched region of the skull. The mice were allowed to recover and were returned to their cages. Three to five weeks later, a 3mm circular craniotomy was performed above the injection site and the craniotomy was sealed with a circular 3mm coverslip with a ~0.5 μm hole to allow pipette access to infected cells. The temperature of the mouse was maintained between 36.5 °C and 37.5 °C throughout the experiment using a homeothermic blanket system (Harvard Instruments). Recordings were of spontaneous activity without visual stimulation. Calcium traces were recorded using a Chameleon Ti-sapphire laser (Coherent) tuned at 920 nm and a 16×, .85 NA Nikon objective. The average power output of the objective was kept < 40 mW. To perform simultaneous loose-patch and two-photon calcium imaging recordings, we used glass pipettes with 7–10 MΩ resistance filled with ACSF and Alexa Fluor 594 (Invitrogen) as described above. Calcium traces were extracted after manually segmenting patched cells and spike times were extracted by thresholding after excluding any periods where the patch was deemed unstable or of low quality. All procedures performed on mice were conducted in accordance with the ethical guidelines of the National Institutes of Health and were approved by the Baylor College of Medicine IACUC.

*Retina*

Imaging experiments were performed as described previously (Briggman and Euler, 2011). In short, the retina was enucleated and dissected from dark-adapted wild-type mice (both genders, C57BL/6J, p21-42), flattened, mounted onto an Anodisc (13, 0.1 mm pores, Whatman) with ganglion cells facing up, and electroporated with Oregon green BAPTA-1 (OGB-1, Invitrogen). The tissue was placed under the microscope, where it was constantly perfused with temperated (36°C) carboxygenated (95% $O_2$, 5% $CO_2$) artificial cerebral spinal fluid (ACSF). Cells were left to recover for at least 1 hour before recordings were performed. We used a MOM-type two-photon microscope equipped with a mode-locked Ti:sapphire laser (MaiTai-HP DeepSee, Newport Spectra-Physics) tuned to 927 nm(Euler et al., 2009). OGB-1 Fluorescence was detected at 520 BP 30 nm (AHF) under a 20x objective (W Plan-Apochromat, 1.0 NA, Zeiss). Data were acquired with custom software (ScanM by M. Müller and T. Euler running under IgorPro 6.3, Wavemetrics), taking 64 x 64 pixel images at 7.8 Hz. Light stimuli were presented through the objective from a DLP projector (K11, Acer), fitted with band-pass-filtered LEDs (amber, z 578 BP 10; and blue/UV, HC 405 BP 10, AHF/Croma), synchronized with the microscope's scanner. Stimulator intensity (as photoisomerization rate, $10^4$ R*/s/cone) was calibrated as described to range from 0.1 (LEDs off) to ~1.3 (ref (Euler et al., 2009)). Mostly due to two-photon excitation of photopigments, an additional, steady illumination component of ~$10^4$ R*/s/cone was present during the recordings. The field of view was 100x100µm, imaging 50-100 cells in the ganglion cell layer simultaneously(Briggman and Euler, 2011). For juxtacellular spike recordings, OGB-1 labeled somata were targeted with a 5 MΩ glass-pipette under dim IR illumination to establish a loose (<1GΩ) seal. Signals were amplified using an Axopatch 200A amplifier (Molecular Devices) in I=0 mode and digitized at 10 kHz on a Digidata 1440A (Molecular Devices). Imaging and spike data were aligned offline using a trigger signal recorded in both acquisition systems, and spike times were extracted by thresholding. All procedures were performed in accordance with the law on animal protection (Tierschutzgesetz) issued by the German Federal Government and were approved by the institutional animal welfare committee of the University of Tübingen.

**Preprocessing**

We normalized the sampling rate of all fluorescence traces and spike trains to 100 Hz, resampling to time bins of 10 ms. This allowed us to apply models across datasets independent of which dataset was used for training. We removed linear trends from the fluorescence traces by fitting a robust linear regression with Gaussian scale mixture residuals. That is, for each fluorescence trace $F_t$, we found parameters $a, b, \pi_k,$ and $\sigma_k$ with maximal likelihood under the model

$$F_t = at + b + \varepsilon_t, \qquad \varepsilon_t \sim \sum_{k=1\ldots K} \pi_k \mathcal{N}(\,\cdot\,; 0, \sigma_k^2),$$

and computed $\widetilde{F}_t = F_t - at - b$. We used three different noise components ($K = 3$). Afterwards, we normalized the traces such that the 5th percentile of each trace's fluorescence distribution is at zero, and the 80[th] percentile is at 1. Normalizing by percentiles instead of the minimum and maximum is more robust to outliers and less dependent on the firing rate of the neuron producing the fluorescence.

**Supervised learning in flexible probabilistic models for spike inference**

We predict the number of spikes $k_t$ falling in the $t$-th time bin of a neuron's spike train based on 1000 ms windows of the fluorescence trace centered around $t$ (preprocessed fluorescence snippets $x_t$). To reduce the risk of overfitting and to speed up the training phase of the algorithm, we reduced the dimensionality of the fluorescence windows via PCA, keeping enough principal components to explain at least 95% of the variance (which resulted in 8 to 20 dimensions, depending on the dataset). Keeping 99% of the variance and slightly regularizing the model's parameters gave similar results but was slower.

We assume that the spike counts $k_t$ given the preprocessed fluorescence snippets $x_t$ can be modeled using a Poisson distribution,

$$p(k_t \mid x_t) = \frac{\lambda(x_t)^k}{k!} e^{-\lambda(x_t)}.$$

We tested three models for the firing rate $\lambda(x_t)$ function:

(1) A spike-triggered mixture (STM) model (Theis et al., 2013) with exponential nonlinearity,

$$\lambda_{\text{STM}}(x_t) = \sum_{k=1}^{K} \exp\left(\sum_{m=1}^{M} \beta_{km}(u_m^\top x_t)^2 + w_k^\top x_t + b_k\right),$$

where $w_k$ are linear filters, $u_m$ are quadratic filters weighted by $\beta_{km}$ for each of $K$ components, and $b_k$ is a offset for each component. We used three components and two quadratic features ($K = 3$, $M = 2$). The performance of the algorithm was not particularly sensitive to the choice of these parameters (we evaluated $K = 1, \ldots 4$ and $M = 1, \ldots, 4$ in a grid search).

(2) As a simpler alternative, we use the linear-nonlinear-Poisson (LNP) neuron with exponential nonlinearity,

$$\lambda_{\text{LNP}}(x_t) = \exp(w^\top x_t + b),$$

where $w$ is a linear filter and $b$ is an offset.

(3) As a more flexible alternative, we used a multi-layer neural network (ML-NN) with two hidden layers,

$$\lambda_{\text{ML-NN}}(x_t) = \exp(w_3^\top g(W_2 g(W_1 x_t + b_1) + b_2) + b_3)$$

,

where $g(y) = \max(0, y)$ is a point-wise rectifying nonlinearity and $W_1$ and $W_2$ are matrices. We tested MLPs with 10 and 5 hidden units, and 5 and 3 hidden units for the first and second hidden layer, respectively. Again, the performance of the algorithm was not particularly sensitive to these parameters.

Parameters of all models were optimized by maximizing the average log-likelihood for a given training set,

$$\frac{1}{N}\sum_{n=1}^{N} \log p(k_t \mid \boldsymbol{x}_t),$$

using limited-memory BFGS (Byrd et al., 1995), a standard quasi-Newton method. To increase robustness against potential local optima in the likelihood of the STM and the ML-NN, we trained four models with randomly initialized parameters and geometrically averaged their predictions. The geometric average of several Poisson distributions again yields a Poisson distribution whose rate parameter is the geometric average of the rate parameters of the individual Poisson distributions.

**Other algorithms**
*SI08*
This approach is based on applying a support-vector machine (SVM) on two PCA features of preprocessed segments of calcium traces. We re-implemented the features following closely the procedures described in (Sasaki et al., 2008). As the prediction signal, we used the distance of the input features to the SVM's separating hyperplane, setting negative predictions to zero. We cross-validated the regularization parameter of the SVM but found that it had little impact on performance.

*PP14*
The algorithm performs Bayesian inference in a generative model, using maximum a posteriori (MAP) estimates for spike inference and MCMC on a portion of the calcium trace for estimating hyperparameters. We used a Matlab implementation provided by the authors of (Pnevmatikakis et al., 2014). We also tried selecting the hyperparameters through cross-validation, which did not substantially change the overall results.

*VP10*
The fast-oopsi or non-negative deconvolution technique constrains the inferred spike rates to be positive (Vogelstein et al., 2010), performing approximate inference in a generative model. We used the implementation provided by the author [1]. We adjusted the hyperparameters using cross-validation by performing a search over a grid of 54 parameter sets controlling the degree of assumed observation noise and the expected number of spikes (Fig. 2a-b). In Fig. 5b-c the hyperparameters were instead directly inferred from the calcium traces by the algorithm.

*YF06*
The deconvolution algorithm (Yaksi and Friedrich, 2006) removes noise by local smoothing and the inverse filter resulting from the calcium transient. We used a Matlab implementation provided by the authors. Using the cross-validation procedure outlined above, we automatically tuned the algorithm by testing 66 different parameter sets. The parameters controlled the cutoff frequency of a low-pass filter, a time constant of the filter used for deconvolution, and whether or not an iterative smoothing procedure was applied to the fluorescence traces.

*OD13*

---
[1] https://github.com/jovo/fast-oopsi

This algorithm performs a template-matching based approach by using the finite rate of innovation-theory as described in (Oñativia et al., 2013). We used the implementation provided on the author's homepage[2]. We adjusted the exponential time constant parameter using cross-validation.

*VP09*

This algorithm performs Bayesian inference in a generative model as described in (Vogelstein et al., 2009). We used the implementation provided by the author[3]. Since this algorithm is based on the same generative model as fast-oopsi but is much slower, we used the hyperparameters inferred by cross-validating fast-oopsi in Fig. 2a-b and the hyperparameters automatically inferred by the algorithm in Fig. 5b-c.

**Performance evaluation**

We evaluated the performance of the algorithms on spike trains binned at 40 ms resolution, i.e. a sampling rate of 25 Hz. For Fig. 3 and Suppl. Fig. 2, we changed the bin width between 10 ms (i.e. 100 Hz) and 500 ms (i.e. 2 Hz). We used cross-validation to evaluate the performance of our framework, i.e. we estimated the parameters of our model on a training set, typically consisting of all but one cell for each dataset, and evaluated its performance on the remaining cell. This procedure was iterated such that each cell was held out as a test cell once. Results obtained using the different training and test sets were subsequently averaged.

*Correlation*

We computed the linear correlation coefficient between the true binned spike train and the inferred one. This is a widely used measure with a simple and intuitive interpretation, taking the overall shape of the spike density function into account. However, the correlation coefficient is invariant under affine transformations, which means that predictions optimized for this measure cannot be directly interpreted as spike counts or firing rates. In further contrast to information gain, it also does not take the uncertainty of the predictions into account. That is, a method which predicts the spike count to be 5 with absolute certainty will be treated the same as a method which experts the spike count to be somewhere between 0 and 10 assigning equal probability to each possible outcome.

*Information gain*

The information gain provides a model based estimate of the amount of information about the spike train extracted from the calcium trace. Unlike AUC and correlation, it takes into account the uncertainty of the prediction.

Assuming an average firing rate of $\lambda$ and a predicted firing rate of $\lambda_t$ at time $t$, the expected information gain (in bits per bin) can be estimated as

$$I_g = \frac{1}{T}\sum_t k_t \log_2 \frac{\lambda_t}{\lambda} + \lambda - \frac{1}{T}\sum_t \lambda_t$$

assuming Poisson statistics and independence of spike counts in different bins. The estimated information gain is bounded from above by the (unknown) amount of information about the spike train contained in the calcium trace, as well as by the marginal entropy of the spike train, which can be estimated using

---
[2] http://www.commsp.ee.ic.ac.uk/%7Epld/software//ca_transient.zip
[3] https://github.com/jovo/smc-oopsi

$$H_m = \frac{1}{T}\sum_t \log(k_t!) - \lambda \log \lambda + \lambda.$$

We computed a relative information gain by dividing the information gain averaged over all cells by the average estimated entropy,

$$\frac{\sum_n I_g^{(n)}}{\sum_n H_m^{(n)}},$$

where $I_g^{(n)}$ is the information gain measured for the $n$-th cell in the dataset.

This can be interpreted as the fraction of entropy in the data explained away by the model (measured in percent points). Since only our method was optimized to yield Poisson firing rates, we allowed all methods a single monotonically increasing nonlinear function, which we optimized to maximize the average information gain over all cells. That is, we evaluated

$$\frac{1}{T}\sum_t k_t \log_2 \frac{f(\lambda_t)}{\lambda} + \lambda - \frac{1}{T}\sum_t f(\lambda_t),$$

where $f$ is a piecewise linear monotonically increasing function optimized to maximize the information gain averaged over all cells (using an SLSQP implementation in SciPy).

*AUC*

The AUC score can be computed as the probability that a randomly picked prediction for a bin containing a spike is larger than a randomly picked prediction for a bin containing no spike (Fawcett, 2006). While this is a commonly used score for evaluating spike inference procedures (Vogelstein et al., 2010), it is not sensitive to changes in the relative height of different parts of the spike density function, as it is invariant under arbitrary strictly monotonically increasing transformations. For example, if predicted rates were squared, high rates would be overproportionally boosted compared to low rates, while yielding equivalent AUC scores.

**Statistical analysis**

We used generalized Loftus & Masson standard errors of the means for repeated measure designs (Franz and Loftus, 2012) and report the mean ± 2 SEM. To assess statistical significance, we compare the performance of the STM model to the performance of its next best competitor, performing a one-sided Wilcoxon signed rank test and report significance or the respective p-value above a line spanning the respective columns. If the STM is not the best model, we perform the comparison between the best model and the STM, coding the comparison in the color of the model.

**Generation of artificial data**

We simulated data by sampling from the generative model used by Vogelstein et al. (2010). That is, we first generated spike counts by independently sampling each bin of a spike train from a Poisson distribution, then convolving the spike train with an exponential kernel to arrive at an artificial calcium concentration, and finally adding Poisson noise to generate a Fluorescence signal $x_t$.

$$k_t \sim \text{Poisson}(\lambda),$$
$$C_t = \gamma C_t + k_t,$$

$$x_t \sim \text{Poisson}(a\, C_t + b).$$

The firing rate $\lambda$ for each cell was randomly chosen to be between 0 and 400 spikes per second. The parameters $\gamma$, $a$, and $b$ were fixed to 0.98, 100 and 1, respectively, and data was generated at a sampling rate of 100 Hz.

**Code and data sharing**

All analysis was done in Python. We provide a Python implementation of our algorithm online (www.bethgelab.org/code/spikeinference)[4].

---

[4] Please note that we are also preparing a Matlab implementation which will be released at a later point in time.

## Acknowledgements

We would like to thank J. Vogelstein, R. Friedrich, E. Pnevtmatikakis and P. Dragotti for making the code for their algorithms available.

This work was supported by the German Federal Ministry of Education and Research (BMBF) through the Bernstein Center for Computational Neuroscience (FKZ 01GQ1002 to T.E., M.B. and A.S.T.); the Deutsche Forschungsgemeinschaft (DFG) through grant BE3848-1 to M.B., BE 5601/1-1 to PB and BA 5283/1-1 to T.B.; the Werner Reichardt Centre for Integrative Neuroscience Tübingen (EXC307); grants NEI R01-EY018847, NEI P30-EY002520-33, and the NIH-Pioneer award DP1-OD008301 to A.S.T.; the McKnight Scholar Award to A.S.T.

## Author contributions

PB, MB and LT designed the project. LT analyzed the data. MF, JR and AST acquired V1 data. MR, TB and TE acquired retinal data. PB wrote the paper with input from all authors. PB and MB supervised the project.

## Table 1: Datasets

| Data set | Area | n | Indicator | Scan frequency | Scanning method | #spikes | sp/s | Field of view |
|---|---|---|---|---|---|---|---|---|
| 1 | V1 | 16 | OGB-1 | 322.5 ± 53.2 | 3D AOD | 19,876 | 1.86 | 200x200 x100 µm³ |
| 2 | V1 | 31 | OGB-1 | 11.8 ± 0.9 | 2D galvo scan | 32,385 | 2.47 | 250x250 µm² |
| 3 | V1 | 19 * (11) | GCamp6s | 59.1 | 2D resonant | 23,974 | 2.58 | 265x265 µm² 135x135 µm² |
| 4 | Retina | 9 | OGB-1 | 7.8 | 2D galvo scan | 12,488 | 4.36 | 100x100 µm² |

* For this dataset, 19 recordings were performed on 11 neurons

## Table 2: Algorithms

| Algorithm | Approach | Technique | Reference |
|---|---|---|---|
| STM | Supervised | STM | This paper |
| SI08 | Supervised | PCA+SVM | (Sasaki et al., 2008) |
| PP14 | Generative | MCMC sampling | (Pnevmatikakis et al., 2014) |
| OD13 | Template matching | Finite rate innovation | (Oñativia et al., 2013) |
| VP10 | Generative | MAP estimation | (Vogelstein et al., 2010) |
| VP09 | Generative | SMC sampling | (Vogelstein et al., 2009) |
| YF06 | Generative | Deconvolution | (Yaksi and Friedrich, 2006) |

# References

Briggman, K.L., and Euler, T. (2011). Bulk electroporation and population calcium imaging in the adult mammalian retina. J. Neurophysiol. *105*, 2601–2609.

Byrd, R.H., Lu, P., Nocedal, J., and Zhu, C. (1995). A Limited Memory Algorithm for Bound Constrained Optimization. SIAM J. Sci. Comput. *16*, 1190–1208.

Chen, T.-W., Wardill, T.J., Sun, Y., Pulver, S.R., Renninger, S.L., Baohan, A., Schreiter, E.R., Kerr, R. a, Orger, M.B., Jayaraman, V., et al. (2013). Ultrasensitive fluorescent proteins for imaging neuronal activity. Nature *499*, 295–300.

Cotton, R.J., Froudarakis, E., Storer, P., Saggau, P., and Tolias, A.S. (2013). Three-dimensional mapping of microcircuit correlation structure. Front. Neural Circuits *7*, 151.

Denk, W., Strickler, J., and Webb, W. (1990). Two-photon laser scanning fluorescence microscopy. Science (80-. ). *248*, 73–76.

Diego, F., and Hamprecht, F.A. (2014). Sparse space-time deconvolution for calcium image analysis. In Neural Information Processing Systems, pp. 1–9.

Euler, T., Hausselt, S.E., Margolis, D.J., Breuninger, T., Castell, X., Detwiler, P.B., and Denk, W. (2009). Eyecup scope--optical recordings of light stimulus-evoked fluorescence signals in the retina. Pflugers Arch. *457*, 1393–1414.

Fawcett, T. (2006). An introduction to ROC analysis. Pattern Recognit. Lett. *27*, 861–874.

Franz, V.H., and Loftus, G.R. (2012). Standard errors and confidence intervals in within-subjects designs: generalizing Loftus and Masson (1994) and avoiding the biases of alternative accounts. Psychon. Bull. Rev. *19*, 395–404.

Froudarakis, E., Berens, P., Ecker, A.S., Cotton, R.J., Sinz, F.H., Yatsenko, D., Saggau, P., Bethge, M., and Tolias, A.S. (2014). Population code in mouse V1 facilitates readout of natural scenes through increased sparseness. Nat. Neurosci. *17*, 851–857.

Greenberg, D.S., Houweling, A.R., and Kerr, J.N.D. (2008). Population imaging of ongoing neuronal activity in the visual cortex of awake rats. Nat. Neurosci. *11*, 749–751.

Grewe, B.F., Langer, D., Kasper, H., Kampa, B.M., and Helmchen, F. (2010). High-speed in vivo calcium imaging reveals neuronal network activity with near-millisecond precision. Nat. Methods *7*, 399–405.

Kerr, J.N.D., and Denk, W. (2008). Imaging in vivo: watching the brain in action. Nat. Rev. Neurosci. *9*, 195–205.

Kerr, J.N.D., Greenberg, D., and Helmchen, F. (2005). Imaging input and output of neocortical networks in vivo. Proc. Natl. Acad. Sci. U. S. A. *102*, 14063–14068.

Lütcke, H., Gerhard, F., Zenke, F., Gerstner, W., and Helmchen, F. (2013). Inference of neuronal network spike dynamics and topology from calcium imaging data. Front. Neural Circuits *7*, 201.

Maruyama, R., Maeda, K., Moroda, H., Kato, I., Inoue, M., Miyakawa, H., and Aonishi, T. (2014). Detecting cells using non-negative matrix factorization on calcium imaging data. Neural Netw. *55*, 11–19.

Oñativia, J., Schultz, S.R., and Dragotti, P.L. (2013). A finite rate of innovation algorithm for fast and accurate spike detection from two-photon calcium imaging. J. Neural Eng. *10*, 046017.

Park, I.J., Bobkov, Y. V., Ache, B.W., and Principe, J.C. (2013). Quantifying bursting neuron activity from calcium signals using blind deconvolution. J. Neurosci. Methods *218*, 196–205.

Pnevmatikakis, E.A., Merel, J., Pakman, A., and Paninski, L. (2013). Bayesian spike inference from calcium imaging data. arXiv /q-bio.NC 0–5.

Pnevmatikakis, E.A., Gao, Y., Soudry, D., Pfau, D., Lacefield, C., Poskanzer, K., Bruno, R., Yuste, R., and Paninski, L. (2014). A structured matrix factorization framework for large scale calcium imaging data analysis. arXiv /q-bio.NC 1–21.

Reimer, J., Froudarakis, E., Cadwell, C.R., Yatsenko, D., Denfield, G.H., and Tolias, A.S. (2014). Pupil Fluctuations Track Fast Switching of Cortical States during Quiet Wakefulness. Neuron *84*, 355–362.

Sasaki, T., Takahashi, N., Matsuki, N., and Ikegaya, Y. (2008). Fast and accurate detection of action potentials from somatic calcium fluctuations. J. Neurophysiol. *100*, 1668–1676.

Stosiek, C., Garaschuk, O., Holthoff, K., and Konnerth, A. (2003). In vivo two-photon calcium imaging of neuronal networks. Proc. Natl. Acad. Sci. U. S. A. *100*, 7319–7324.

Theis, L., Chagas, A.M., Arnstein, D., Schwarz, C., and Bethge, M. (2013). Beyond GLMs: a generative mixture modeling approach to neural system identification. PLoS Comput. Biol. *9*, e1003356.

Valmianski, I., Shih, A.Y., Driscoll, J.D., Matthews, D.W., Freund, Y., and Kleinfeld, D. (2010). Automatic identification of fluorescently labeled brain cells for rapid functional imaging. J. Neurophysiol. *104*, 1803–1811.

Vogelstein, J.T., Watson, B.O., Packer, A.M., Yuste, R., Jedynak, B., and Paninski, L. (2009). Spike inference from calcium imaging using sequential Monte Carlo methods. Biophys. J. *97*, 636–655.

Vogelstein, J.T., Packer, A.M., Machado, T. a, Sippy, T., Babadi, B., Yuste, R., and Paninski, L. (2010). Fast nonnegative deconvolution for spike train inference from population calcium imaging. J. Neurophysiol. *104*, 3691–3704.

Wilt, B. a, Fitzgerald, J.E., and Schnitzer, M.J. (2013). Photon shot noise limits on optical detection of neuronal spikes and estimation of spike timing. Biophys. J. *104*, 51–62.

Yaksi, E., and Friedrich, R.W. (2006). Reconstruction of firing rate changes across neuronal populations by temporally deconvolved Ca2+ imaging. Nat. Methods *3*, 377–383.

# Figure captions

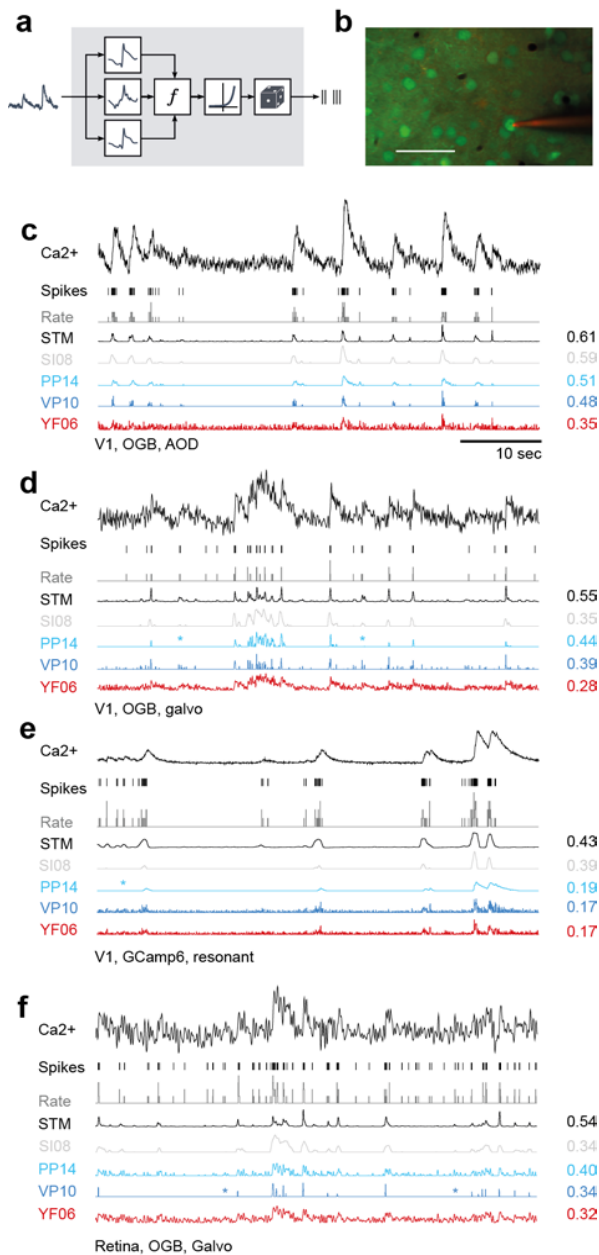

**Figure 1: Spike inference from calcium measurements**

a) Schematic of the probabilistic STM model.
b) Simultaneous recording of spikes and calcium fluorescence traces in primary visual cortex of anesthetized mice. Green: Cells labeled with OGB-1 indicator. Red: Patch pipette filled with Alexa Fluor 594. Scale bar: 50 μm.
c) Example cell recorded from V1 using AOD scanner and OGB-1 as indicator. From top to bottom: Calcium fluorescence trace, spikes, spike rate at 25 Hz (grey), inferred spike rate using the STM model (black), SI08, PP14, VP14 and YF06. All traces were scaled independently for clarity. On the right, correlation between the inferred and the original spike rate is shown.
d) Example cell recorded from V1 using galvanometric scanner and OGB-1 as indicator. For legend, see c).
e) Example cell recorded from V1 using resonance scanner and GCamp6s as indicator. Note the different indicator dynamics. For legend, see c).
f) Example cell recorded from the retina using galvanometric scanner and OGB-1 as indicator. For legend, see c).

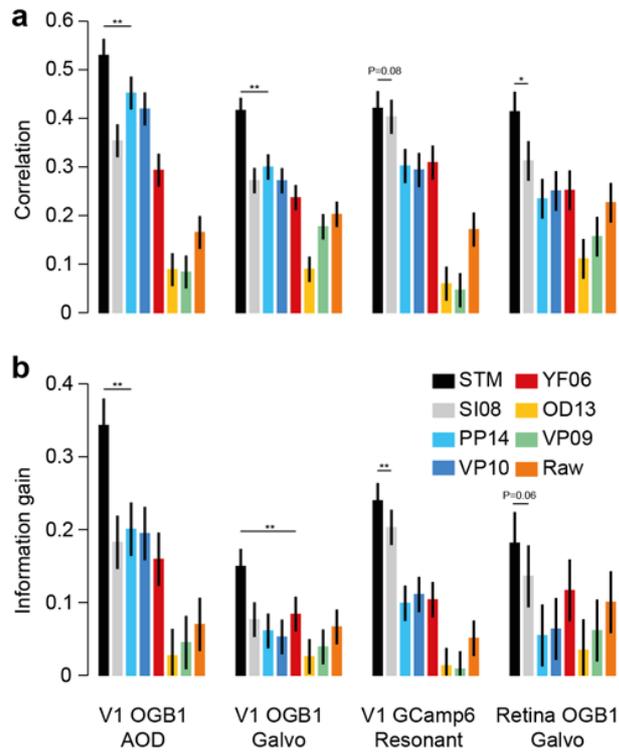

**Figure 2: Quantitative evaluation of spike inference performance**
a) Correlation (mean± 2 SEM for repeated measure designs) between the true spike rate and the inferred spike density function for different algorithms (see legend for color code) evaluated on the four different datasets (with n=16, 31, 19 and 9, respectively). Markers above bars show the result of a Wilcoxon sign rank test between the STM model and its closest competitor (see *Methods*, * denotes P<0.05, ** denotes P<0.01).
b) Information gained about the true spike train by observing the calcium trace, evaluated for different algorithms.

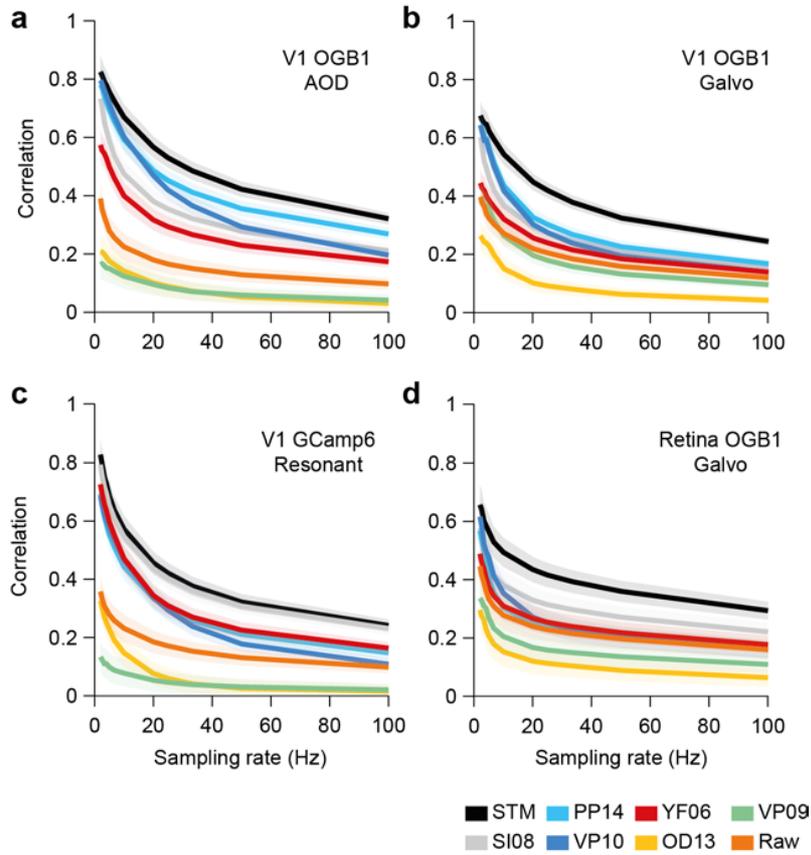

**Figure 3: Spike inference performance as a function of frequency**
Correlation (mean ± 2 SEM for repeated measure designs) between the true and inferred spike rate as a function of frequency for all four datasets (a-d) with n=16, 31,19 and 9, respectively. See legend for color code.

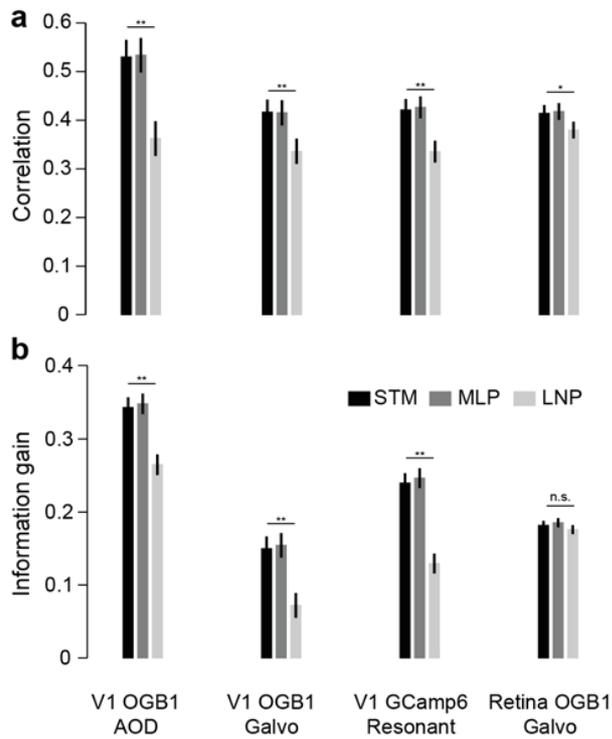

**Figure 4: Evaluating model complexity**
a) Correlation (mean ± 2 SEM for repeated measure designs) between the true and inferred spike rate comparing the STM model (black) with a flexible multilayer neural network (dark grey) and a simple LNP model (light grey) evaluated on the four different datasets (with n=16, 31, 19 and 9, respectively). Markers above bars show the result of a Wilcoxon sign rank test between the STM model and the LNP model (see *Methods*, * denotes P<0.05, ** denotes P<0.01).
b) Information gained about the true spike train by observing the calcium trace performing the same model comparison described in a).

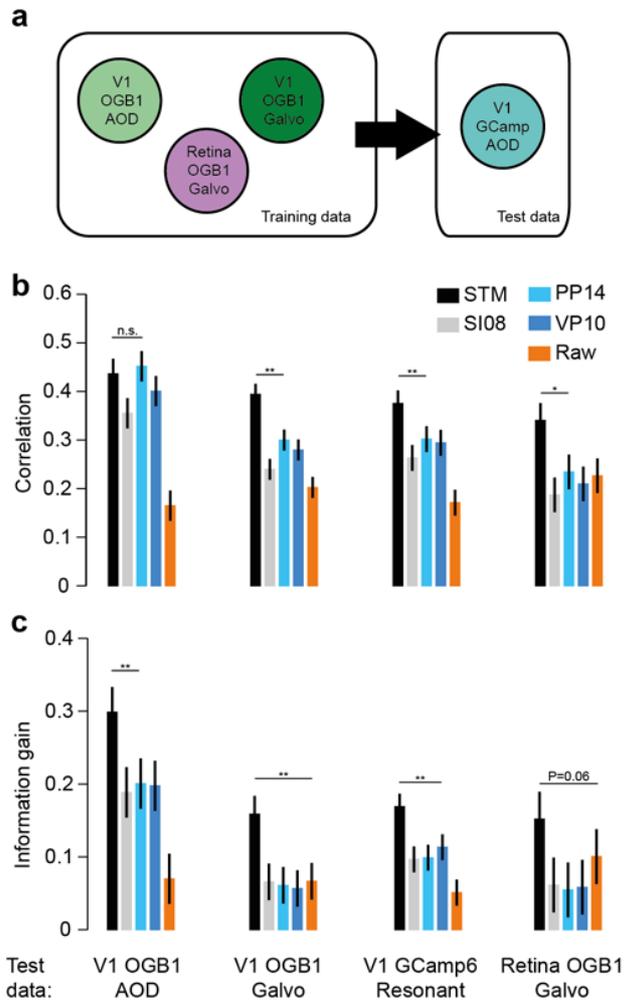

**Figure 5: Testing generalization performance**

a) Schematic illustrating the generalization setup: The algorithms are trained on all cells from three datasets (here: all but the GCamp dataset) and evaluated on the remaining dataset (here: the GCamp dataset), testing how well its performance carries over to datasets it has not seen during training.

b) Correlation (mean± 2 SEM for repeated measure designs) between the true spike rate and the inferred spike density function for a subset of the algorithms (see legend for color code) evaluated on each of the four different datasets (with n=16, 31, 19 and 9, respectively), trained on the remaining three. Markers above bars show the result of a Wilcoxon sign rank test between the STM model and its closest competitor (see *Methods*, * denotes P<0.05, ** denotes P<0.01).

c) Information gained about the true spike train by observing the calcium trace performing the generalization analysis described in a).

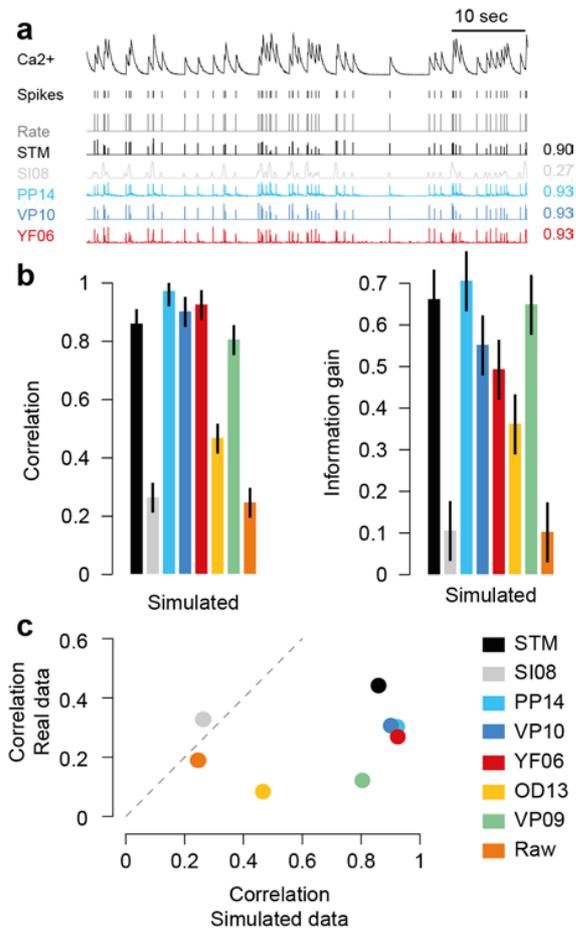

**Figure 6: Evaluating algorithms on artificial data**

a) Example trace sampled from a generative model, true spikes and binned rate as well as reconstructed spike rate from four different algorithms (conventions as in Fig. 1). Numbers on the right denote correlations between true and inferred spike trains.

b) Correlation (mean± 2 SEM for repeated measure designs) and information gain computed on a simulated dataset with 20 traces. For algorithms see legend.

c) Scatter plot comparing performance on simulated data with that on real data (averaged over cells from all datasets), suggesting little predictive value of performance on simulated data.

# Supervised learning sets benchmark for robust spike detection from calcium imaging signals

Lucas Theis[1,2]*, Philipp Berens[$*,1,2,3,4,5], Emmanouil Froudarakis[4], Jacob Reimer[4], Miroslav Román Rosón[1,5], Tom Baden[1,3,5], Thomas Euler[1,3,5], Andreas Tolias[3,4,6], Matthias Bethge[1,2,3]

[1] Centre for Integrative Neuroscience, University of Tübingen, Tübingen, Germany
[2] Institute of Theoretical Physics, University of Tübingen, Tübingen, Germany
[3] Bernstein Center for Computational Neuroscience, University of Tübingen, Tübingen, Germany
[4] Department of Neuroscience, Baylor College of Medicine, Houston, USA
[5] Institute for Ophthalmic Research, University of Tübingen, Tübingen, Germany
[6] Department of Computational and Applied Mathematics, Rice University, Houston, USA

* These authors contributed equally to this work.

$ To whom correspondence should be addressed:
Philipp Berens, philipp.berens@uni-tuebingen.de

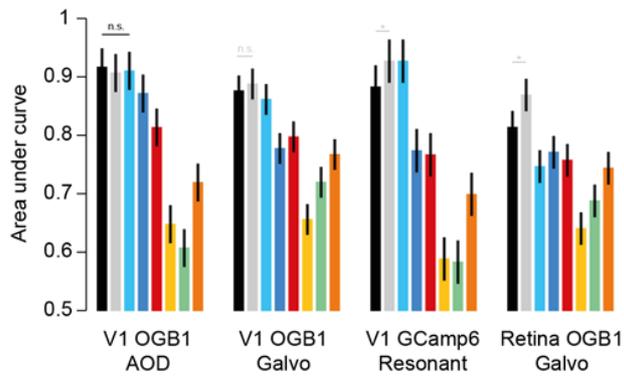

**Supplementary Figure 1 (related to Fig. 2)**
AUC (mean± 2 SEM for repeated measure designs) between the true spike rate and the inferred spike density function for different algorithms (see legend for color code) evaluated on the four different datasets (with n=16, 31, 19 and 9, respectively). Markers above bars show the result of a Wilcoxon sign rank test between the STM model and its closest competitor (see Methods, * denotes $P<0.05$, ** denotes $P<0.01$).

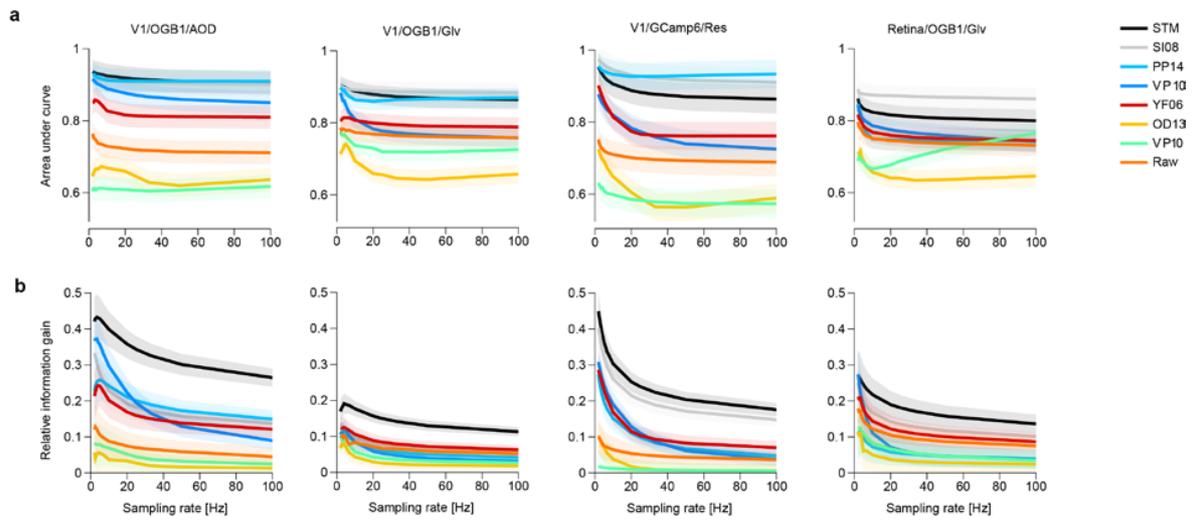

**Supplementary Figure 2 (related to Fig. 3)**

a) Information gain (mean ± 2 SEM for repeated measure designs) between the true and inferred spike rate as a function of frequency for all four datasets (a-d) with n=16, 31,19 and 9, respectively. See legend for color code.
b) AUC, as above.

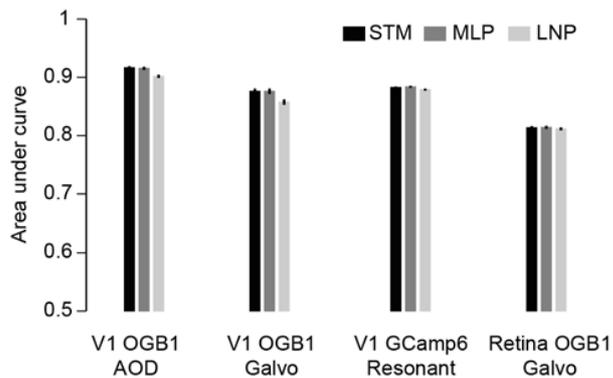

**Supplementary Figure 3 (related to Fig. 4)**
AUC (mean ± 2 SEM for repeated measure designs) between the true and inferred spike rate comparing the STM model (black) with a flexible multilayer neural network (dark grey) and a simple LNP model (light grey) evaluated on the four different datasets (with n=16, 31, 19 and 9, respectively). Markers above bars show the result of a Wilcoxon sign rank test between the STM model and the LNP model (see Methods, * denotes P<0.05, ** denotes P<0.01).

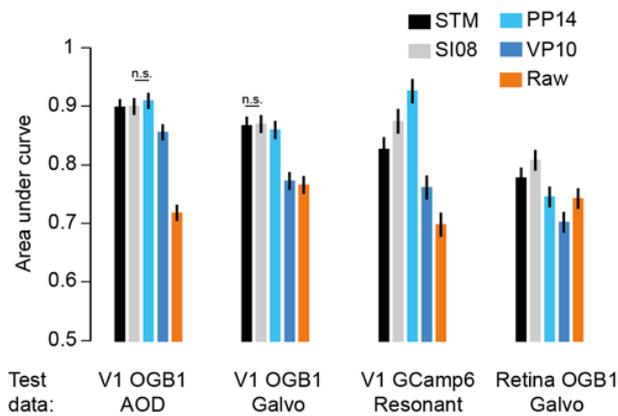

**Supplementary Figure 4 (related to Fig. 5)**
AUC (mean± 2 SEM for repeated measure designs) between the true spike rate and the inferred spike density function for a subset of the algorithms (see legend for color code) evaluated on each of the four different datasets (with n=16, 31, 19 and 9, respectively), trained on the remaining three. Markers above bars show the result of a Wilcoxon sign rank test between the STM model and its closest competitor (see Methods, * denotes P<0.05, ** denotes P<0.01).

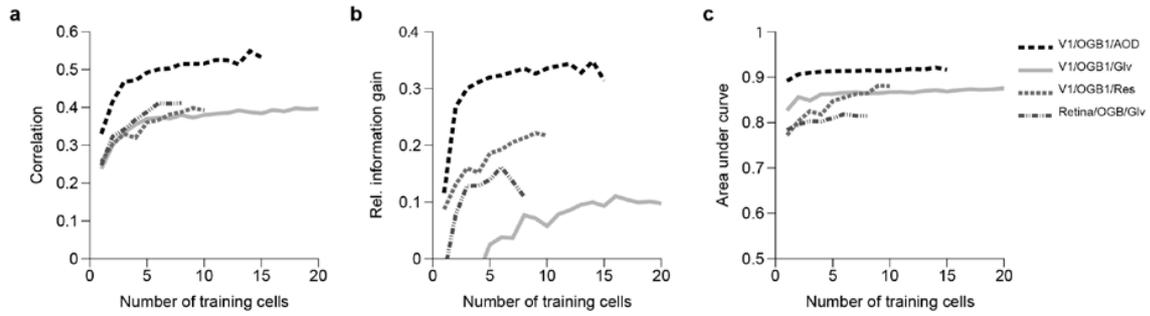

**Supplementary Figure 5: Training set size**
  a) Correlation as a function training set size for the four different datasets (see legend). Error bars are omitted for clarity.
  b) As in a), but for Information gain.
  c) As in a), but for AUC.